%% file: main.tex
\documentclass[letterpaper, 10 pt, conference]{ieeeconf}  

\IEEEoverridecommandlockouts                              

\usepackage{graphicx} 
\usepackage{multirow}
\usepackage{booktabs}
\usepackage{pifont}
\usepackage{amsmath} 
\usepackage{amssymb}  
\usepackage{mathtools}
\usepackage{cite}

\usepackage{hyperref}

\usepackage{color}
\usepackage{xcolor}

\usepackage[abs]{overpic}
\usepackage{pict2e}

\newcommand{\R}{\mathbb{R}}
\newcommand{\p}{\text{p}}
\newcommand{\cmark}{\ding{51}}%
\newcommand{\xmark}{\ding{55}}%

\title{\LARGE \bf
Multi-view object pose estimation from \\
correspondence distributions and epipolar geometry
}

\author{Rasmus Laurvig Haugaard and Thorbj{\o}rn Mosekj{\ae}r Iversen 
\thanks{
All authors are from SDU Robotics, Maersk Mc-Kinney Moller Institute, University of Southern Denmark.
The authors gratefully acknowledge the support from Innovation Fund Denmark through the project MADE Fast.
\newline
{\tt\small \{rlha,thmi\}@mmmi.sdu.dk}}
}

\begin{document}

\maketitle
\thispagestyle{plain}
\pagestyle{plain}

\input{tex/abstract}
\input{tex/introduction}

\input{tex/relatedWork}
\input{tex/method}
\input{tex/evaluation}
\input{tex/results}
\input{tex/limitations}
\input{tex/conclusion}





{\small
\bibliographystyle{IEEEtran}
\bibliography{references}
}

\end{document}

%% file: tex/abstract.tex
\begin{abstract}
In many automation tasks involving manipulation of rigid objects, the poses of the objects must be acquired. 
Vision-based pose estimation using a single RGB or \mbox{RGB-D} sensor is especially popular due to its broad applicability.
However, single-view pose estimation is inherently limited by depth ambiguity and ambiguities imposed by various phenomena like occlusion, self-occlusion, reflections, etc.
Aggregation of information from multiple views can potentially resolve these ambiguities, but the current state-of-the-art multi-view pose estimation method only uses multiple views to aggregate single-view pose estimates, and thus \textit{rely} on obtaining good single-view estimates. 
We present a multi-view pose estimation method which aggregates learned 2D-3D distributions from multiple views for both the initial estimate and optional refinement. 
Our method performs probabilistic sampling of 3D-3D correspondences under epipolar constraints using learned 2D-3D correspondence distributions which are implicitly trained to respect visual ambiguities such as symmetry.
Evaluation on the T-LESS dataset shows that our method reduces pose estimation errors by 80-91\% compared to the best single-view method, and we present state-of-the-art results on T-LESS with four views, even compared with methods using five and eight views.
\end{abstract}

%% file: tex/introduction.tex
\section{INTRODUCTION}
\label{sec:introduction}

\begin{figure}[thpb]
    \centering
    \begin{overpic}[width=\columnwidth,unit=1mm]{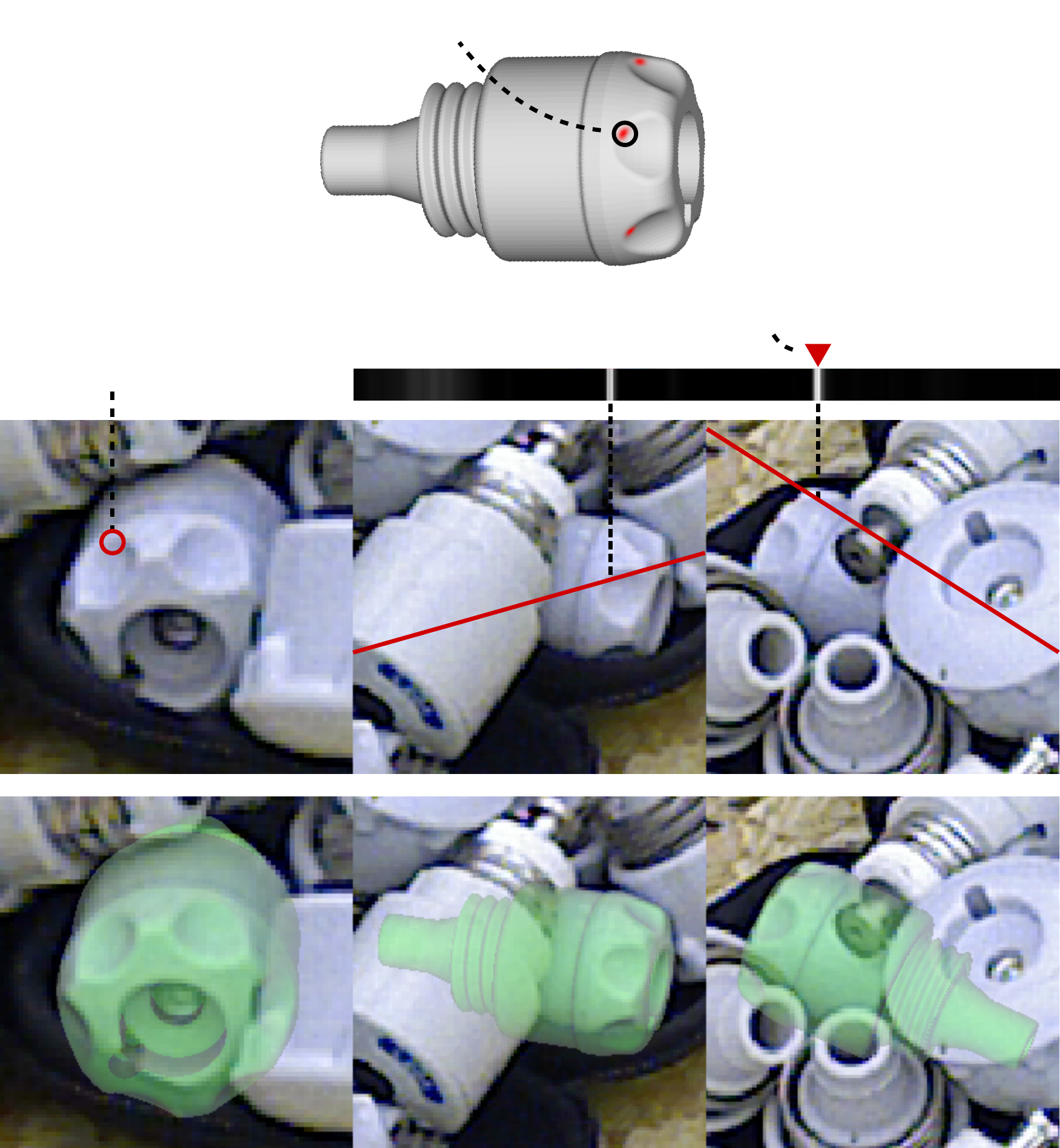}
        \put(3.5,63.5){1) $u_1\sim \p(u_1)$}
        \put(32,92){2) $c\sim \p(c|u_1)$}
        \put(57,68){3) $u_2\sim \p(u_2|u_1, c)$}
    \end{overpic}
    \caption{
        A real example of sampling a 3D-3D correspondence using three image crops from different views.
        We propose to sample from the joint 3D-3D correspondence distribution $\p(x, c)$ of scene points, $x\in \R^3$, and object points, $c\in \R^3$.
        We let $x$ be represented by two points, $(u_1, u_2)$, from two different views, such that $\p(x, c) = \p(u_1,u_2,c) = \p(u_1)\p(c|u_1)\p(u_2|u_1,c)$.
        %
        1) An image point, $u_1$, shown by a red circle, is sampled from estimated masks across views. For brevity, the masks are not shown here.
        2) An object point, $c$, shown by a black circle, is sampled from a learned 2D-3D distribution over the object's surface. The distribution $\p(c|u_1)$ is is shown in red on the 3D model.
        3) The image point, $u_1$, imposes epipolar lines in the other images, shown by red lines.
        We use the learned 2D-3D distribution as well as the epipolar constraints to approximate $\p(u_2|u_1,c)$, which is shown in black and white, and sample $u_2$ from this distribution, shown by a red triangle.
        The resulting pose estimate from our full pipeline is superimposed in the bottom row. 
    }
    \label{fig:epipolar}
\end{figure}

Many robotics tasks involve precise manipulation of rigid objects. This is especially true for industrial robotics where high precision pose estimates are crucial for successful execution of demanding tasks such as bin picking and assembly. 
Single-view pose estimation applies to a wide range of scenarios where pose estimation is desired, from augmented reality to self-driving cars to object manipulation, including bin picking and assembly.
Therefore, from both a practical and scientific perspective, it is of great interest to push the limits of single-view pose estimation.
For example, much effort has been focused on obtaining robustness to partial occlusions, 
with popular benchmarks targeting pose estimation even for object that are 90\% occluded \cite{hodavn2020bop}, 
and thus, state-of-the-art methods show quite remarkable robustness towards occlusion.
Single-view pose estimation is however inherently limited, not only by occlusion from other objects, but also by self-occlusion and notably depth ambiguity.
Estimating depth from a single color image has relatively high uncertainty as a consequence of the small effect change of depth can have on the size of the object in the image. 
Depth sensors can be used to reduce depth ambiguity,
however, 
the quality of cheap depth sensors is questionable, 
good depth sensors are expensive,
and it does not mitigate the other ambiguities from single-view pose estimation.

In some cases, like industrial object manipulation, obtaining multiple views \textit{is} feasible, and intuitively, aggregation of information from multiple views has great potential to reduce ambiguities and obtain more robust systems.
State-of-the-art multi-view pose estimation methods either only use multi-view information for refinement \cite{labbe2020cosypose, dpodv2} or use heuristic features \cite{de2022cendernet}. 
Also, \cite{labbe2020cosypose, dpodv2, de2022cendernet} assume uni-modal distributions in pose space, 2D-3D correspondence space, and curvature space, respectively.


We hypothesize that there is potential in taking an approach that is more fundamentally multi-view,
while using learned features that can represent relevant ambiguities.
To that end, we envision a pipeline consisting of a multi-view detector followed by a multi-view pose estimator, and this work focuses on the latter.

We present a novel multi-view RGB-only pose estimation method, 
using recent learned 2D-3D correspondence distributions \cite{surfemb} and epipolar geometry to sample 3D-3D correspondences as outlined in Figure~\ref{fig:epipolar}.
Pose hypotheses are sampled from the resulting correspondences, and scored by the agreement with the 2D-3D correspondence distributions across views.
Finally, the best pose hypothesis is refined to maximize this agreement across views.


\medskip

Our primary contribution is a state-of-the-art multi-view RGB-only pose estimation method, combining learned 2D-3D correspondence distributions and epipolar geometry.
We also contribute multi-view extensions of the pose scoring and refinement proposed in \cite{surfemb}.

\medskip

We present related work in \ref{sec:relatedwork}, describe our method in detail in \ref{sec:method}, how we evaluate our method in \ref{sec:evaluation}, present and discuss our findings in \ref{sec:results}, and comment on limitations and future work in \ref{sec:limitations}.

%% file: tex/relatedWork.tex
\section{RELATED WORK}
\label{sec:relatedwork}
The following section reviews the literature related to the problem of estimating 6D poses of rigid objects from multiple RGB or RGB-D sensors, under the assumption that accurate 3D models of the objects are available. 
Even under this limited scope, there exists a large number of published methods, and thus this review is limited to recent publications which serves as representative works of the various pose estimation methodologies.

Using a course categorization, pose estimation methods can be divided into surface-based methods and image-based methods. Surface-based methods rely on the creation of a 3D point cloud by reconstructing points of an object's surface, e.g. using traditional stereo methods which triangulates surface points from 2D-2D point correspondences under epipolar constrains~\cite{hartley2003multiple}, using deep learning based multi-view stereo~\cite{yao2018mvsnet}, or depth sensors~\cite{grenzdorffer2020ycb}. The depth information from multiple RGB-D views have also been fused to form more complete point clouds~\cite{zeng2017multi}. The object pose can then be inferred from the point cloud by finding 3D-3D correspondences between the cloud and the model. This has e.g. been done using deep learning to estimate 3D object keypoints directly from sparse point clouds~\cite{kaskman20206}, or estimating 2D-3D correspondences and lifting them to 3D-3D correspondences with a depth sensor \cite{shugurov2021dpodv2}. From 3D-3D correspondences, the Kabsch algorithm can be used to compute the relative pose, often as part of a RANSAC procedure~\cite{fischler1981random} to be robust toward outliers. Pose refinement is then often performed, e.g. using ICP~\cite{besl1992method}.

Surface-based methods rely on the accuracy of the reconstructed point clouds, so their performance is dependent the quality of the depth sensor or that accurate 2D-2D correspondences can be found for triangulation. This can be problematic for industrial applications since accurate depth sensors are expensive and industrial objects tend to be symmetric with featureless surfaces, which makes for a challenging correspondence problem~\cite{hodavn2020bop}.

Image-based pose estimation methods estimate a pose directly from the image without reconstructing the object surface in 3D. One approach to image-based pose estimation is establishing 2D-3D correspondences followed by the Perspective-n-Point (PnP) algorithm~\cite{fischler1981random}, which utilizes that the projection of 4 or more non-colinear object points uniquely define an object's pose. The PnP algorithm is the foundation for many both traditional and contemporary pose estimation methods, e.g. \cite{rad2017bb8} which uses deep learning to regress bounding box corners, and \cite{surfemb} learns 2D-3D correspondence distributions. Other image-based deep learning methods include direct regression of object orientation~\cite{labbe2020cosypose,Xiang2018posecnn} and treating the rotation estimation as a classification problem\cite{kehl2017ssd}.

While most pose estimation methods assume a uni-modal pose distribution and handle symmetries explicitly to better justify this assumption, such as \cite{labbe2020cosypose, rad2017bb8}, there are methods such as \cite{surfemb}, which allows multi-modal pose distributions and handles ambiguities like symmetry implicitly.

All of the previously mentioned methods focus on estimating \textit{the} single most probable pose, given the image. There also exists methods that estimate entire pose distributions~\cite{murphy2021implicit,kipode2022iversen}.

The above image-based pose estimation methods are all single-view, and thus suffer from the aforementioned single-view ambiguities. Several pose estimation methods have been proposed to aggregate information from multiple views. These can roughly be divided into one of two categories depending on whether the aggregation is done as high level pose refinement or low level image feature aggregation. 

High-level view aggregation has e.g. been done using object-level SLAM which simultaneously refines the poses of multiple objects together with multi-view camera extrinsics~\cite{fu2021multi}, pose voting which increases reliability of pose estimates through a voting scheme~\cite{li2018unified}, or pose refinement by minimizing a loss based on similarity between observed and rendered correspondence features across views~\cite{shugurov2021dpodv2,shugurov2021multi}. 
Most of the methods assume that the object symmetries are provided, such that the pose ambiguities caused by object symmetry can be explicitly accounted for. 
This has e.g. been done by explicitly incorporating symmetries in the objective of a scene graph optimization~\cite{labbe2020cosypose}, or using symmetries together with estimated extrinsics to explicitly add priors on the object pose~\cite{merrill2022symmetry}.
Methods that estimate full pose distributions \cite{murphy2021implicit,kipode2022iversen} are particularly well suited for pose-level multi-view aggregation~\cite{naik2022multi}.
However, the methods have yet to show state-of-the-art performance on established benchmarks.

Low-level aggregation of image data from multiple views has been done by using DenseFusion~\cite{wang2019densefusion} to aggregate learned visual and geometric features from multiple views~\cite{duffhauss2022MV6D}, by estimating centroids and image curvatures for use in a multi-view render-and-compare scheme~\cite{de2022cendernet}, or by formulating pose refinement as an optimization problem with an objective based on the object shape reprojection error in multiple views\cite{li2022bcot}.

Current state-of-the-art multi-view pose estimation \cite{labbe2020cosypose} is achieved by enforcing consistency among single-view pose information across multiple views.
This method does not assume that the camera extrinsics are available, 
and jointly refines camera extrinsics and object poses from single-view pose estimates in a bundle-adjustment.
Since \cite{labbe2020cosypose} fuses multi-view information from single-view pose estimates, they rely on at least two of the initial single-view estimates to be accurate, they do not benefit from views with poor single-view pose estimates, and they rely on explicit symmetries. 
In contrast, our method utilizes all views through the entire pose estimation pipeline, and handles ambiguities like symmetries implicitly.

To the best of our knowledge, our work is the first to show probabilistic sampling of 3D-3D correspondences using 2D-3D correspondence distributions and epipolar geometry.

%% file: tex/method.tex
\section{METHOD}
\label{sec:method}
This section will first motivate and provide an overview of our method. 
The following subsections will provide more detailed information on the individual parts.

Let $x \in \R^3$ be a 3D point in the scene's frame, and $c \in R^3$ be a 3D point in the object's frame.
A true 3D-3D correspondence, $(x, c)$, denotes the same point in the two frames, and from three such correspondences, the object's pose can be found with the Kabsch algorithm.
Let $I \in \R^{H \times W \times 3}$ be an image crop, and let $V$ denote a set of image crops of the same object instance from different views.
To obtain 3D-3D correspondences towards pose estimation, we aim to sample from the joint conditional distribution 
\begin{equation}
    \p(x, c | V).
\end{equation}
For brevity, conditioning on $V$ is assumed in the following.
A 3D scene point $x$ can be uniquely determined by triangulation from two compatible image points, $(u_1, u_2)$, $u \in \R^2$, from different views, and thus $\p(x, c) = \p(u_1, u_2, c)$.
Factorizing the joint distribution, we obtain
\begin{equation}
    \p(x, c) = \p(u_1, u_2, c) = \p(u_1) \p(c|u_1) \p(u_2|u_1, c),
    \label{eq:expand-joint}
\end{equation}
which represents the general approach of this work, also shown in Figure~\ref{fig:epipolar}.
To sample from $\p(x, c)$, we
first sample an image point, $u_1$, across all views from estimated masks, $\p(u_1)$.
Then, we sample an object point, $c$, from an estimated 2D-3D correspondence distribution, $\p(c|u_1)$.
Lastly, we sample an image point in another view, $u_2$, from an estimated 3D-2D distribution with epipolar constraints, $\p(u_2|u_1, c)$.

As mentioned, the object point, $x$, in the 3D-3D correspondence, $(x, c)$, is simply found by triangulating $(u_1, u_2)$, and a pose estimate is obtained from a triplet of such correspondences using Kabsch.
Ambiguities in $\p(x, c)$ may lead to many incoherent correspondence triplets, which motivates filtering triplets based on geometric constraints.
Finally, poses from geometrically valid triplets are scored by a multi-view pose score, and the best scoring pose is refined to maximize this score.


\subsection{Sampling from \texorpdfstring{$\p(u_1|V)$}{p(u1|V)}}
\label{sec:sampling-2d}

First, we need to properly define the desired distribution of 3D-3D correspondences, $\p(x, c | V)$.
We define $\p(x, c | V)$ by assigning equal probability to the 3D-3D correspondences across pixels within the mask across views, from which it follows that
\begin{equation}
    \p(u_1|V) \propto \p(u_1\in M | V),
\end{equation}
where $M$ is the set of pixels within the true masks across views.  
Then, assuming that there is little to no benefit from extending mask-estimation to multi-view, that is
$\p(u_1 \in M | V) = \p(u_1 \in M | I_1)$,
where $I_1$ denotes the image from which $u_1$ belongs,
it follows that
\begin{equation}
\label{eq:sampling-2d}
    \p(u_1 | V) \propto \p(u_1 \in M | I_1),
\end{equation}
and we thus sample $u_1$ proportional to estimated single-view mask probabilities.

Note that while Eq.~\ref{eq:sampling-2d} is an assumption, single-view mask estimation is not affected by self-occlusion, depth-ambiguity or symmetry, and occlusion leads to ambiguities along hidden boundaries, which do not \textit{prevent} sampling of correct 3D-3D correspondences, although it may increase the amount of outliers.

\subsection{Revisiting SurfEmb's 2D-3D distributions}
SurfEmb~\cite{surfemb} consists of an encoder-decoder convolutional neural network, $f$, referred to as the query model, that maps a color image, $I \in \R^{H \times W \times 3}$, to a query embedding image, $Q \in \R^{H \times W \times E}$, as well as a fully connected neural network, $g$, referred to as the key model, mapping 3D object coordinates, $c\in\R^3$, to key embeddings, $k\in\R^E$.
Together, the two models represent a 2D-3D correspondence distribution, 
\begin{equation}
    \p(c_i | u, I) \propto \exp(q_u^T k_i),
\end{equation}
where $q_u = f(I)_{u}$ is the query embedding at image point $u$, and $k_i = g(c_i)$ is the key embedding for the object coordinate $c_i$.
The models are trained jointly with a contrastive loss to maximize the probability of the true correspondence, implicitly learning to represent the correspondence distributions with respect to visual ambiguities like symmetry.



\subsection{Sampling from \texorpdfstring{$\p(c|u_1, V)$}{p(c|u1,V)}}
\label{sec:sampling-2d-3d}
Similar as the assumption in Section~\ref{sec:sampling-2d}, 
we make the assumption that the individual 2D-3D distributions do not benefit significantly from having access to the other views, 
\begin{equation}
    \p(c|u_1, V) = \p(c|u_1, I_1).
\end{equation}
We thus sample from $\p(c|u_1, I_1)$ as estimated by SurfEmb.

Similar to the discussion of single-view mask estimation in \ref{sec:sampling-2d}, single-view 2D-3D correspondence distributions does not suffer from depth-ambiguity, and more views do not resolve ambiguities related to global symmetries. 

\subsection{Sampling from \texorpdfstring{$\p(u_2|u_1, c, V)$}{p(u2|u1,c,V)}}
\label{sec:sampling-epi}

\begin{figure}
    \centering
    \includegraphics[width=\columnwidth]{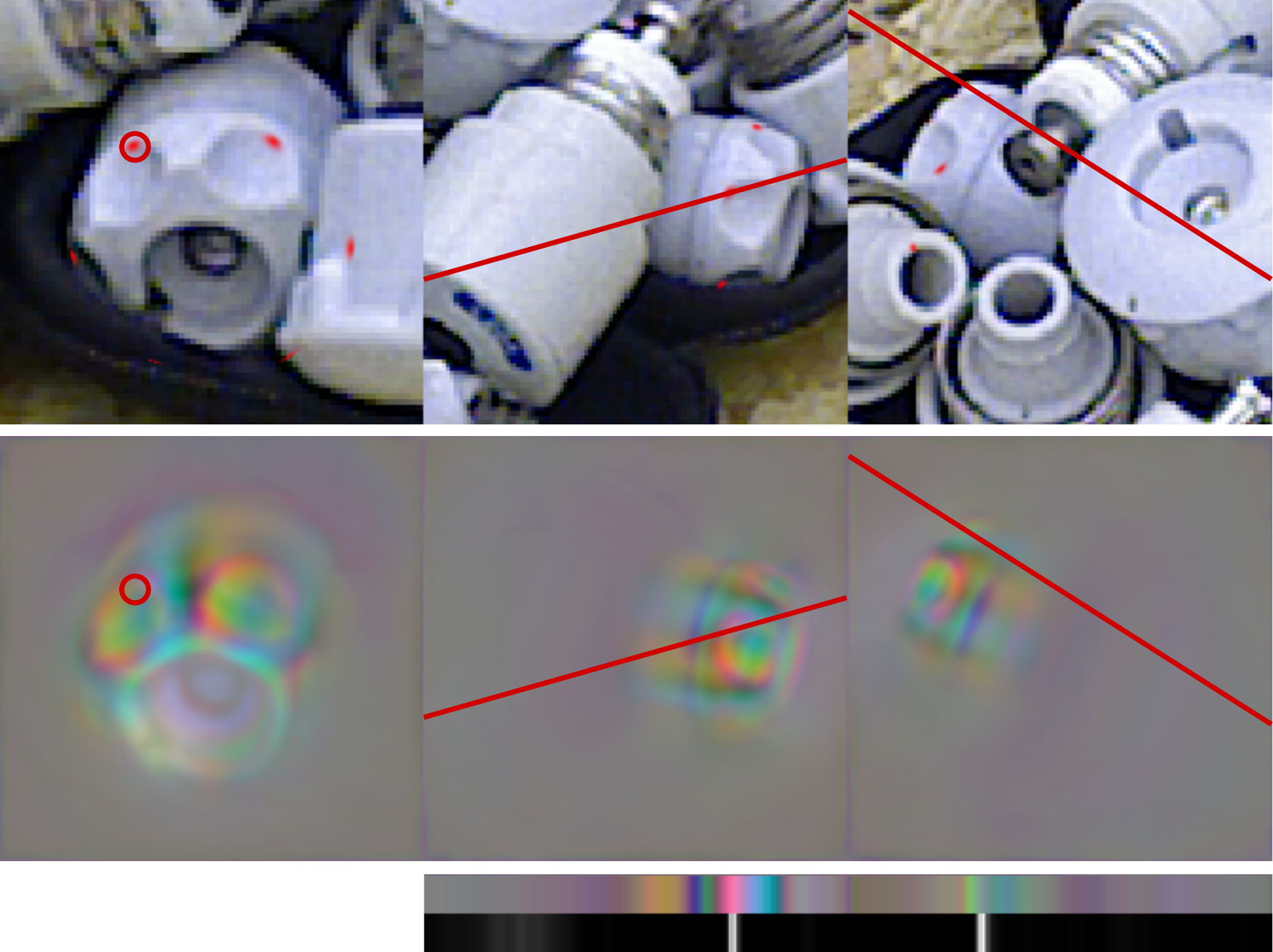}
    \caption{
        Query embedding visualizations for the 3D-3D correspondence sample in Fig.~\ref{fig:epipolar}.
        The image crops in the first row are passed through SurfEmb's query model, which provides query embedding images, shown in the second row.
        The image point $u_1$, shown by a red circle, imposes epipolar constraints in the other images, shown by red lines.
        The query embeddings along the epipolar lines are shown in the last row together with the resulting distribution $\p(u_2|u_1,c)$.
        The 2D-3D correspondence distribution, $\p(c|u_1)$, is superimposed on the first row of images in red.
    }
    \label{fig:embeddings}
\end{figure}

We assume conditioning on $V$ in the following to enhance readability.
From bayes rule, it follows that
\begin{equation}
\label{eq:u2-bayes}
    \p(u_2|u_1, c) =
    \frac{
        \p(u_2) \p(c | u_2) \p(u_1| c, u_2) 
    }{
        \p(u_1, c)
    }.
\end{equation}

$\p(u_2|u_1, c)$ is normalized over $u_2$, eliminating the constant denominator, $\p(u_1, c)$.
We approximate $\p(u_2|V) \propto \p(u_2\in M|I_2)$ as in Section~\ref{sec:sampling-2d},
and approximate $\p(c | u_2, V) = \p(c | u_2, I_2)$, as in Section~\ref{sec:sampling-2d-3d},
which leaves $\p(u_1|c, u_2, V)$ to be determined.

We approximate $\p(u_1|c, u_2, V)$ using epipolar geometry by setting $\p(u_1|c, u_2, V) \propto 1$ for all point pairs $(u_1, u_2)$ that share the same epipolar plane, and zero otherwise. 
Note that this is a conservative approximation in the sense that it is more ambiguous than the true distribution, $\p(u_1|c, u_2, V)$, but $u_1$ and $u_2$ must, by definition, lie on the same epipolar plane.

To summarize, we approximate Eq.~\ref{eq:u2-bayes} as
\begin{equation}
\label{eq:epi-approx}
    \p(u_2|u_1, c) \propto \p(u_2\in M|I_2) \p(c | u_2, I_2),
\end{equation}
normalizing over $u_2$ along the epipolar lines in the other images imposed by $u_1$.
See Fig.~\ref{fig:embeddings}.
Note that $\p(u_1|c, u_2)$ disappeared because Eq. \ref{eq:epi-approx} is only considered along the epipolar lines, on which we defined $\p(u_1|c, u_2)$ to be uniform.

Finally, we find the 3D scene point, $x$, by triangulating $(u_1, u_2)$, which provides the 3D-3D correspondence, $(x, c)$.

\subsection{Sampling pose hypotheses}
\label{sec:sampling-pose}

We use the Kabsch algorithm to establish pose hypotheses from uniformly sampled triplets of 3D-3D correspondences.
Ambiguities like symmetry as well as the previous assumptions can cause a large ratio of incoherent triplets, even when the independent correspondences in a triplet are correct up to symmetry.
Because of this and because the pose scoring described in Section~\ref{sec:scoring-refinement} is relatively computationally expensive, we score and filter the triplets by estimating a geometric signal to noise ratio, similar to \cite{buch2013pose}, as described next.

A triplet describes two triangles. One in the scene frame, from the points found by triangulation, $(x_1, x_2, x_3)$, and one in the object frame by the points sampled from the object surface, $(c_1, c_2, c_3)$.
For a perfectly accurate triplet of correspondences, the two triangles are equal up to a rigid transformation: the object's pose.
In that case, it follows that the side lengths of the triangles are equal, $||c_i - c_j|| = ||x_i - x_j||$.
We let the L2 norm of differences between the three side lengths represent the triplet's noise.

The sensitivity of the pose found by Kabsch with respect to noise depends on the triangle. A small triangle or a triangle whose points are close to colinear are more sensitive to noise.
We let the smallest triangle height, across baselines and the two triangles, represent the signal.
We thus score the triplets with respect to the ratio between the minimum triangle height and side length differences, and then only perform Kabsch and pose scoring on the best triplets.

\subsection{Multi-view pose scoring and pose refinement}
\label{sec:scoring-refinement}
We implement straight forward multi-view extensions of SurfEmb's single-view pose scoring and pose refinement, by averaging across views.
For completeness, this section summarizes the single-view versions.
We refer to \cite{surfemb} for details.

The single-view pose score used in SurfEmb consists of two parts.
A mask score and a correspondence score. 
The mask score is defined as the estimated average log likelihood of the mask, $\hat M$, of a pose estimate,
\begin{equation}
    s_M = \frac{1}{|U|} \sum_{u\in U} \log \p(\hat M_u | I),
\label{eq:mask-score}
\end{equation}
where $U$ is the set of all pixels in the image. 
The correspondence score is defined as the estimated average log-likelihood of the object coordinates within the mask of the estimated pose,
\begin{equation}
    s_C = \frac{1}{|\hat M|} \sum_{u\in \hat M} \log \p(c_u|u, I).
\label{eq:corr-score}
\end{equation}
The total pose score is a weighted sum of the scores, normalized by maximum entropy,
\begin{equation}
    s = \frac{s_M}{\log(2)} + \frac{s_C}{\log(|\hat S|)},
\end{equation}
where $|\hat S|$ is the number of object points in the 2D-3D distribution.
Pose refinement is done by a local maximization of the correspondence score (Eq.~\ref{eq:corr-score}) with respect to the pose estimate, using only the visible object coordinates at the initial pose estimate.
The Broyden–Fletcher–Goldfarb–Shanno (BFGS) algorithm is used for the optimization.

As mentioned, we simply average over all views to obtain multi-view versions of the pose score and the pose refinement objective.
In \cite{surfemb}, the motivation to only refine the pose using the correspondence score was that the mask score was not easily differentiable. 
It was attempted to include the part of the mask score from within the mask of the pose estimate, however, they observed that with single-view refinement, this lead to an objective that was biased towards pushing objects away from the camera.
With multi-view constraints however, we have not observed this to be the case, and so we include this partial mask score in the refinement objective.

\subsection{Implementation details}
Establishing the 2D-3D correspondence distribution for a query is $O(|\hat S|E)$, where $|\hat S|$ is the number of object coordinates and $E$ is the number of embedding dimensions, while the 3D-2D distribution is $O(VWE)$, where $V$ is the number of views and $W$ is the number of pixels along each epipolar line.
In our case, establishing 2D-3D correspondences is significantly more expensive than establishing the 3D-2D distribution, $VW \approx 1.000 << |\hat S| \approx 70.000$. Also, sampling $m$ samples from the distributions is $O(N + m\log N)$, where $N$ is $VW$ and $|\hat S|$, respectively.
We can thus speed up sampling significantly by sampling more than one object coordinate per 2D-3D distribution and more than one pixel-coordinate per 3D-2D distribution.

In practice, we sample 5000 samples from $p(u_1)$, five samples from $p(c|u_1)$ per $u_1$, and five samples from $p(u_2|u_1, c)$ per $(u_1, c)$, providing a total of $125.000$ correspondences. 
In constrast to sampling once per distribution, we obtain 25x more 3D-3D correspondence, in less than 2x the compute.

We uniformly sample 125.000 triplets from the sampled 3D-3D correspondences,
take the 1000 best triplets based on the estimated signal-to-noise ratio,
run multi-view pose scoring on those and refine the best scoring pose.
The time required for a pose estimate is approximately 1-2 seconds with the above parameters.

%% file: tex/evaluation.tex
\section{EVALUATION}
\label{sec:evaluation}
We aim to design experiments that answer the following questions:
\begin{enumerate}
    \item How much improvement is our multi-view method able to show over the best single-view method?
    \label{question:multi-vs-single}
    \item How does our method compare to other multi-view methods?
    \label{question:multi-vs-multi}
\end{enumerate}

\newcommand{\tabdata}[9]{& #1 & #4 & #2 & #3 & #6 & #5 & #7 & #8 & #9 \\}
\begin{table*}
    \centering
    \small
    \caption{
        Pose estimation results on T-LESS. 
        GT: ground truth detections.
        RGB: the method only uses color images, and thus requires no depth sensor.
        PBR: the method is only trained on synthetic renders and thus requires no annotated real images.
    }
    \resizebox{\linewidth}{!}{%
    \begin{tabular}{llccc|cc|ccc}
    Detections%
        \tabdata{Method}{RGB}{PBR}{\#views}{$\uparrow$ AR}{$\downarrow$ 1-AR}{AR\textsubscript{VSD}}{AR\textsubscript{MSSD}}{AR\textsubscript{MSPD}}
    \toprule
    CenDerNet \cite{de2022cendernet} \tabdata{CenDerNet \cite{de2022cendernet}}{\cmark}{\cmark}{5}{0.713}{0.287}{0.707}{0.717}{0.715}
    YOLO~\cite{redmon2018yolov3} \tabdata{DPODv2 \cite{dpodv2}}{\cmark}{\xmark}{4}{0.720}{0.280}{0.679}{0.742}{0.740} 
    \cline{1-1}
    \multirow{8}{*}{MaskRCNN~\cite{he2017mask}}%
        \tabdata{CosyPose \cite{labbe2020cosypose}}{\cmark}{\cmark}{1}{0.640}{0.360}{0.571}{0.589}{0.761}
        \tabdata{CosyPose \cite{labbe2020cosypose}}{\cmark}{\xmark}{1}{0.728}{0.272}{0.669}{0.695}{0.821}
        \tabdata{\quad + ICP}{\xmark}{\xmark}{1}{0.701}{0.299}{0.587}{0.749}{0.767}
        \tabdata{\quad + Multi-view \cite{labbe2020cosypose}}{\cmark}{\xmark}{4}{0.801}{0.199}{0.742}{0.795}{0.864}
        \tabdata{\quad + Multi-view \cite{labbe2020cosypose}}{\cmark}{\xmark}{8}{0.839}{0.161}{0.773}{0.836}{0.907}
        \tabdata{SurfEmb \cite{surfemb}}{\cmark}{\cmark}{1}{0.735}{0.265}{0.661}{0.686}{0.857}
        \tabdata{\quad + Depth ref. \cite{surfemb}}{\xmark}{\cmark}{1}{0.828}{0.172}{0.797}{0.829}{0.859}
        \tabdata{\quad\quad + Multi-view ref. (ours)}{\xmark}{\cmark}{4}{\textbf{0.858}}{\textbf{0.142}}{\textbf{0.843}}{\textbf{0.863}}{\textbf{0.869}}
    \midrule
    \multirow{4}{*}{GT}%
        \tabdata{DPODv2 \cite{dpodv2}}{\cmark}{\cmark}{4}{0.795}{0.205}{-}{-}{-}
        \tabdata{SurfEmb \cite{surfemb}}{\cmark}{\cmark}{1}{0.843}{0.157}{0.764}{0.794}{0.969}
        \tabdata{\quad + Depth ref. \cite{surfemb}}{\xmark}{\cmark}{1}{0.929}{0.071}{0.891}{0.928}{0.970}
        \tabdata{\quad\quad + Multi-view ref. (ours)}{\xmark}{\cmark}{4}{\textbf{0.959}}{\textbf{0.041}}{\textbf{0.937}}{\textbf{0.964}}{\textbf{0.976}}
    \midrule
    \multirow{2}{*}{Multi-view GT}%
        \tabdata{EpiSurfEmb (ours)}{\cmark}{\cmark}{4}{0.976}{0.024}{0.944}{0.990}{0.994}
        \tabdata{\quad + Multi-view ref. (ours)}{\cmark}{\cmark}{4}{\textbf{0.986}}{\textbf{0.014}}{\textbf{0.965}}{\textbf{0.994}}{\textbf{0.998}}
    \bottomrule
    \end{tabular}%
    }
    \vspace{1.4em}
    \label{tab:main_results}
\end{table*}

As will be clear in Section~\ref{sec:results}, for state-of-the-art single-view results, approximately half of the errors can be attributed to poor detections.
To isolate the performance of pose estimation, we evaluate our method on ground truth crops similar to DPODv2~\cite{dpodv2}.

We evaluate on the T-LESS~\cite{tless} dataset, with the metrics defined in the BOP benchmark~\cite{hodavn2020bop}.
BOP defines three pose errors: 
Visible Surface Discrepancy (VSD),
Maximum Symmetry-Aware Surface Distance (MSSD), and
Maximum Symmetry-Aware Projection Distance (MSPD),
and calculates the recall at different thresholds to obtain an average recall over thresholds for each error.
The total average recall, $AR$, is then the average recall over the three metrics.
We refer to \cite{hodavn2020bop} for details about the pose errors.

We follow other multi-view methods \cite{labbe2020cosypose, dpodv2} and provide results with four views.
We use the same sets of views as in CosyPose~\cite{labbe2020cosypose}.
Note that the T-LESS test dataset in the BOP Benchmark has 20 scenes, each with 50 views, which makes 12 sets of 4 views and 1 set of only 2 views.
Because of this and because not all instances are visible from all views, the ground truth detections have 28, 192, 281 and 1292 sets of 1, 2, 3 and 4 image crops respectively, in total in the test set.
For the 28 single-view image crops, we use the single-view SurfEmb method based on 2D-3D correspondences and PnP, otherwise we use our multi-view method based on 3D-3D correspondences and Kabsch.

To best answer question~\ref{question:multi-vs-single}, we compare SurfEmb's single view method with our multi-view method, both with ground truth crops to isolate errors from detection.
To further clarify how much of the improvement comes from multi-view pose hypotheses and multi-view refinement, we also apply our multi-view refinement on single-view estimates.

Answering question~\ref{question:multi-vs-multi} is not as straight forward.
Ideally, we could compare the multi-view parts of methods in isolation, but the multi-view parts have different assumptions and thus rely on the rest of the respective methods, and so it's not clear how to properly approach the question this way.
For example, our method uses multi-view information both to obtain pose hypotheses and to refine them, and so it's not directly comparable to methods that solely focus on multi-view pose refinement.
Another problem is that our method assumes multi-view detections (linked crops), and it would not for example be fair to CosyPose~\cite{labbe2020cosypose} to enforce the knowledge about linked crops and apply their multi-view refinement across all linked views, since a bad pose in one view would reduce their average recall across views.
It is also not fair to apply our multi-view refinement on CosyPose pose estimates, since CosyPose predicts poses up to the provided symmetries, even when the objects are not actually visually symmetric, and thus may not provide a good initial guess for multi-view refinement based on SurfEmb's learned distributions.
Note that CosyPose jointly estimates object poses and extrinsics, however they state that their method does not benefit significantly from knowing the true extrinsics.

Instead of comparing multi-view parts of methods in isolation, we believe the best way to answer question~\ref{question:multi-vs-multi} is to compare the methods in full.
Since to the best of our knowledge, 
no other method has presented results on ground truth multi-view detections, 
and since we are not aware of any off-the-shelf multi-view detector,
we are limited to compare our multi-view refinement with other multi-view methods.
To this end, we use SurfEmb's single-view pose estimates with depth refinement to extract linked image crops in all views. 

%% file: tex/results.tex
\section{RESULTS}
\label{sec:results}

We present our results in Table~\ref{tab:main_results}.
To better convey our results, we show 1-AR, which is the on average missing recall, which we will simply refer to as the error.
The first thing to notice is the impact of detections on single-view results. 
Ground truth detections reduce the SurfEmb error by 41-59\%, before and after depth refinement, respectively, which means approximately half of the errors can be attributed to poor detections. This motivates using ground truth crops to compare pose estimation in isolation.

Our method with ground truth multi-view detections reduces the error by 80-91\% compared to the best single-view method with ground truth detections, depending on whether depth is available for SurfEmb or not.
Also, the results show that while applying our multi-view refinement to the single-view estimates do show a 42\% error reduction, our full method has a further 66\% error reduction.
This answers question~\ref{question:multi-vs-single}.

SurfEmb's single-view method with our multi-view pose refinement has an 80\% error reduction compared to DPODv2's multi-view approach on ground truth crops.
However, note that SurfEmb's single-view method is already better than DPODv2.
Perhaps more interestingly, we also show that SurfEmb's single-view estimates on real detections with depth refinement followed by our multi-view refinement is state-of-the-art on T-LESS~\cite{tless}.
This partially answers question~\ref{question:multi-vs-multi}.

%% file: tex/limitations.tex
\section{LIMITATIONS AND FUTURE WORK}
\label{sec:limitations}

Because our method uses the learned SurfEmb~\cite{surfemb} embeddings, 
we inherit its limitations.
We thus require relatively good CAD models, as \cite{surfemb} showed that the embeddings can have trouble generalizing to real images in case of baked in lighting in the 3D model textures.

SurfEmb only models the 2D-3D distribution for parts of the model that are not self-occluded.
If we also modeled the distribution for self-occluded parts of the object, the probabilities from multiple views could be aggregated elegantly to form a field of 3D-3D correspondence distributions.
However, expanding the SurfEmb model to include self-occluded parts is non-trivial. 
Currently, a query only needs to contain enough information to express the 2D-3D correspondence distribution related to the nearest object surface point along the queries ray.
This is not only an easier than estimating which self-occluded surface points the ray may hit. 
It also limits the \textit{variety} of 2D-3D distributions which motivates using a relatively low-dimensional embedding space. Extending the 2D-3D distributions to also represent self-occluded object points, we're asking the embeddings to express distributions describing all the combination of surface points that can lies on a ray. 
From an information theory perspective, this greatly increases the entropy of the distribution over 2D-3D distributions, $\mathrm{H}(p( p(c|I, u)))$, which might render the capacity of the embedding space insufficient and make it harder to generalize. 
A further exploration of this challenge is left for future work.




%% file: tex/conclusion.tex
\section{CONCLUSION}
\label{sec:conclusion}
We have proposed a way to utilize recent 2D-3D correspondence distributions combined with epipolar geometry to sample 3D-3D correspondences, 
given image crops of the same object instance from multiple views.
We also proposed a way to sample and prune 3D-3D correspondence triplets to obtain pose hypotheses.
Furthermore, we propose a multi-view pose scoring method and a multi-view pose refinement method.
Our results on ground truth detections showed that our full pipeline reduces the pose estimation errors by 80-91\% compared to the best single-view method.
With real detections, our multi-view refinement with four views sets a new state-of-the-art on T-LESS, even compared with methods using five and eight views.